\documentclass[11pt]{article}

\usepackage[final]{acl}

\usepackage{times}
\usepackage{latexsym}

\usepackage[T1]{fontenc}
\usepackage[utf8]{inputenc}

\usepackage{microtype}

\usepackage{inconsolata}

\usepackage{graphicx}
\usepackage{amsmath}
\usepackage{multirow}
\usepackage{makecell}
\usepackage{subcaption}
\usepackage{xcolor}

\title{WISE-Flow: Workflow-Induced Structured Experience for Self-Evolving Conversational Service Agents}

\author{
Yuqing Zhou$^{\dagger1}$\quad Zhuoer Wang$^2$ \quad Jie Yuan$^2$ \quad Hong Wang$^2$ \quad Samson Koelle$^2$ \\
\textbf{Ziwei Zhu$^1$\quad Wei Niu$^2$} \\
$^1$George Mason University \quad $^2$Amazon\\
\texttt{\{yzhou31, zzhu20\}@gmu.edu, \{wze, niuwei\}@amazon.com}
}

\begin{document}
\maketitle

\renewcommand{\arraystretch}{1.2}
\renewcommand{\thefootnote}{\fnsymbol{footnote}}
\footnotetext[2]{The research was done during an internship at Amazon.}
\renewcommand{\thefootnote}{\arabic{footnote}}

\begin{abstract}
Large language model (LLM)-based agents are widely deployed in user-facing services but remain error-prone in new tasks, tend to repeat the same failure patterns, and show substantial run-to-run variability. Fixing failures via environment-specific training or manual patching is costly and hard to scale. To enable self-evolving agents in user-facing service environments, we propose WISE-Flow, a workflow-centric framework that converts historical service interactions into reusable procedural experience by inducing workflows with prerequisite-augmented action blocks. At deployment, WISE-Flow aligns the agent's execution trajectory to retrieved workflows and performs prerequisite-aware feasibility reasoning to achieve state-grounded next actions. Experiments on ToolSandbox and $\tau^2$-bench show consistent improvement across base models.

\end{abstract}

\section{Introduction}
By augmenting large language models (LLMs) with external tools and an iterative action-observation loop, LLM-based agents can resolve multi-step tasks beyond text generation~\cite{lewis2020retrieval, schick2023toolformer, yao2022react}. Such agents are increasingly deployed for task-oriented user-facing services. However, they can still make mistakes in practice, especially on new tasks or unfamiliar environments, repeatedly fail in similar ways, and exhibit high variance across runs. Fixing these failures via environment-specific training or manual patching is costly and hard to scale. This motivates self-evolving agents that improve over time by themselves~\cite{gao2025survey}.

A key aspect of self-evolution is the ability to learn from past experiences~\cite{gao2025survey}. Recent work has studied memory-augmented agents that store experience from prior interactions and retrieve it to guide future decisions. Abstract summary memories capture high-level summaries or reflections~\cite{zhong2024memorybank, shinn2023reflexion}. While they provide general takeaways, they often remain \textbf{free-form and abstract}, providing limited support for deriving concrete, executable procedures, especially in the early stages of a task. Interaction trace memories store fine-grained traces, such as individual tool calls or events~\cite{zhang2023large, xu2025mem}. Although useful for recalling specific past steps, these traces are often \textbf{fragmented and lack procedural structure}, e.g., an explicit task-level workflow with prerequisites and recovery paths. Distillation methods further improve usability by converting offline trajectories into compact guidelines ~\cite{fu2024autoguide} or workflows~\cite{wang2024agent}. However, distilled guidelines are often \textbf{too coarse to align with the agent’s live progress} and yield the next feasible action under the current tool state. Even when workflow memories include step-level state descriptions, they typically \textbf{do not encode action prerequisites} as explicit, checkable constraints for execution-time validation. Collectively, these limitations leave a persistent retrieval-to-action gap. 
%A complete discussion of related work is in Appendix~\ref{app:related_work}. 

To address the aforementioned gaps, we propose Workflow-Induced Structured Experience (WISE-Flow), a workflow-centric framework for self-evolving agents, which turns past experiences into reusable procedural experience rather than isolated steps or free-form reference. WISE-Flow provides execution-time guidance for conversational agents in user-facing services. This guidance takes the form of workflows containing action prerequisites. Specifically, WISE-Flow induces \textbf{workflows together with prerequisite-augmented action blocks}, carrying both a task-level procedural backbone and the conditions under which actions tend to be valid. During the deployment, WISE-Flow \textbf{aligns the ongoing interaction trace to the retrieved workflows and performs prerequisite-aware feasibility checks}, turning retrieved experience into stage-appropriate next-step guidance that is grounded in the agent's current world state, rather than leaving the model to infer the procedure and feasibility from scratch.

In addition to the framework, we improve agent evaluation. Prior work emphasizes end-to-end success (e.g., task success rate)~\cite{barres2025tau, lu2025toolsandbox} but overlooks procedural action errors during execution, which also affect the user experience. We therefore adopt the concept of $F_\beta$~\cite{Chinchor1992MUC4EM} and adapt it to service-agent evaluation to jointly measure milestone progress and error control, offering additional insights beyond the final outcome.

In summary, our contributions are as follows: (1) We propose a two-phase framework, workflow-induced structured experience (WISE-Flow), that automatically induces workflows and prerequisite-augmented action blocks from raw interaction logs and integrates them into a generic tool-augmented agent via workflow-level and action-level retrieval. (2) We introduce a new workflow representation and an evaluation metric $F_\beta$ to jointly reflect milestone completion and error control. (3) We demonstrate the gains of WISE-Flow across benchmarks and base models. Specifically, it improves $F_\beta$ by at least 3\% on $\tau^2$-bench over the strongest baseline and achieves the highest success rate.

\begin{figure*}
    \centering
    \includegraphics[width=0.9\linewidth]{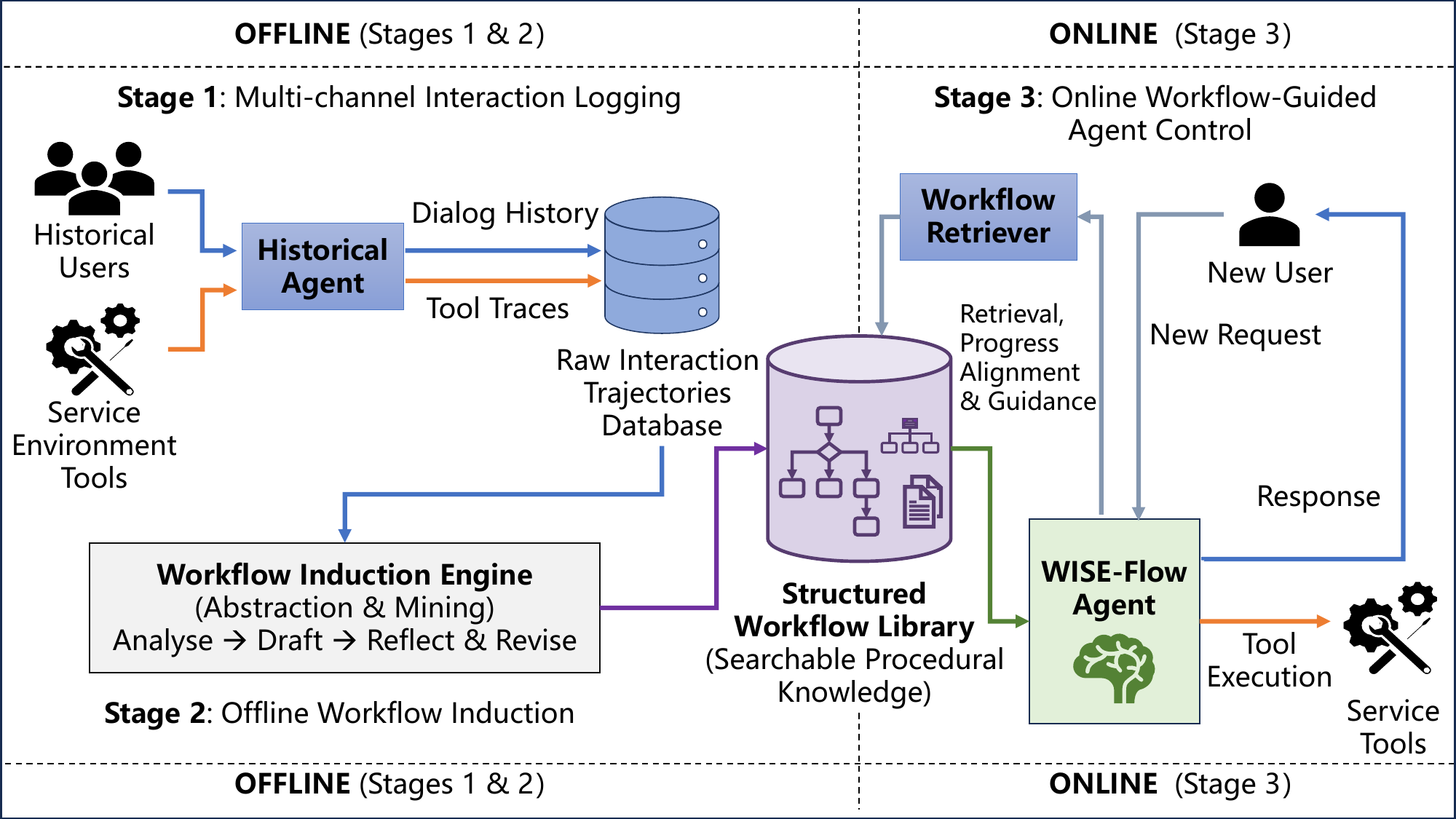}
    \caption{WISE-Flow pipeline.}
    \vspace{-11pt}
    \label{fig:pipeline}
\end{figure*}

\section{Related Work}
\label{app:related_work}
\paragraph{Tool-augmented conversational agents and benchmarks.} Tool-augmented conversational agents extend LLMs by enabling them to interact with external tools and environments, providing access to external information and executable capabilities beyond the models' static knowledge~\cite{lewis2020retrieval, schick2023toolformer, qin2023toolllm, mialon2023augmented}. They often follow an iterative action–observation loop, where tool calls produce environment feedback that guides subsequent decisions~\cite{nakano2021webgpt, yao2022react}. Building on this paradigm, prior work has studied both tool orchestration and improving tool invocation reliability through training and large-scale tool-use supervision~\cite{patil2024gorilla, schick2023toolformer}. In user-facing services, tool use is typically stateful and procedural. Later actions depend on intermediate results, backend state, and coordination with the user~\cite{yao2024tau, lu2025toolsandbox, barres2025tau}. Accordingly, recent benchmarks increasingly evaluate agents under evolving world states and long-horizon interactions, such as $\tau$-bench, ToolSandbox, and $\tau^2$-bench~\cite{yao2024tau, lu2025toolsandbox, barres2025tau}. These settings motivate our focus on execution-time guidance, i.e., reusing past interaction experience as procedural knowledge to guide the next action under the current world state.

\paragraph{Self-evolving agents and memory-augmented experience reuse.} To improve reliability in long-running deployments, an emerging direction is to build self-evolving agents that improve over time from past interaction experience~\cite{gao2025survey, wang2023voyager}. A common approach is to equip agents with explicit memory, storing past interactions and retrieving them to inform future decisions~\cite{gao2025survey, zhang2025survey, zhao2024expel, park2023generative, chhikara2025mem0, salama2025meminsight}. In conversational settings, some methods maintain high-level summaries or reflective memories that capture general takeaways from past dialogues~\cite{zhong2024memorybank, shinn2023reflexion}. Other methods store fine-grained traces, such as tool calls, events, or state-action records, enabling retrieval of locally relevant steps~\cite{zhang2023large, xu2025mem}. While these memory forms support experience reuse, they often leave open how to represent and apply past experience as \emph{procedural} guidance during execution in stateful tool environments, especially when the agent must localize progress and select a feasible next action under the current world state.

\paragraph{Procedural experience.} Recent work seeks to turn interaction trajectories into reusable procedural knowledge for tool execution. A first line produces textual procedural guidance, distilling experience into artifacts such as conditional guidelines~\cite{fu2024autoguide}, concise retrievable hints~\cite{nekoei2025just}, or manual-like instructions~\cite{chen2024automanual} that can be consulted during planning. These memories are easy to store and reuse, but they are usually not tied to the agent's exact stage in an ongoing execution and often leave the applicability of actions implicit. A second direction focuses on structured procedural representations~\cite{wang2024agent, fang2025memp}. While such representations capture stepwise procedural knowledge, they are typically retrieved and injected as the task-level context, without explicitly aligning the current interaction trace to a procedure step or validating next-step feasibility at execution time. In contrast, WISE-Flow induces workflow with prerequisite-augmented action blocks from interaction logs and bridges retrieval to action via progress alignment and prerequisite-aware feasibility checks, turning retrieved experience into stage-appropriate and feasible next-step guidance at execution time. Moreover, most procedural-reuse methods are studied in settings that tolerate exploratory trial-and-error during execution, such as web navigation~\cite{zhou2023webarena} and embodied interactive simulators~\cite{shridhar2020alfworld}. In contrast, we focus on user-facing services, where procedural failures directly degrade user experience. Therefore, agents should minimize intermediate errors, not just achieve end-to-end success.

\section{Methodology}
In this section, we introduce WISE-Flow, a pipeline for tool-augmented conversational agents operating in task-oriented service environments. The agent's goal is to resolve the user's request as efficiently as possible, with minimal steps and errors. To achieve the objective, we treat historical interaction trajectories as reusable procedural knowledge. Specifically, we extract structured workflows from past interactions and organize them into a searchable library. At inference time, the agent retrieves workflows relevant to the new task and uses them as actionable guidance and soft constraints when planning its next actions.

At a high level, WISE-Flow follows a three-stage pipeline as shown in Figure~\ref{fig:pipeline}: (1) collect full logs from multiple sources, including dialog, tool traces and environment feedback, (2) automatically induce structured workflows from the collected logs, and (3) guide agents using the induced workflows. Next, we describe each stage in turn.

\subsection{Multi-channel Interaction Logging}
The first stage builds a repository of service logs. In task-oriented services, many failures are procedural rather than linguistic, such as executing actions in an invalid order or violating implicit prerequisites. Such failures are often only revealed through environment-side feedback, and are difficult to diagnose from dialog text alone. Consequently, dialog-only summaries are insufficient to recover \emph{reusable procedures}. 
Unlike previous work~\cite{choubey-etal-2025-turning}, WISE-Flow explicitly emphasizes the importance of modeling the \emph{complete service log}, not just a sequence of utterances, but a sequence of events from multiple sources. The richer record can provide the evidence needed to recover reliable procedures offline for future reference. Formally, each trajectory is modeled as an ordered event stream: $\tau = \{e_t\}_{t=1}^T,\ e_t = (r_t, x_t)$, where $r_t\in \{user, assistant, environment\}$ means the source of the event, and $x_t$ is its content, which can be user messages, assistant responses, tool invocations, tool outputs, and backend signals such as API return codes and error messages.

\subsection{Offline Workflow Induction}
Given a set of service trajectories $\mathcal{T}$ collected for a task, the second stage of WISE-Flow performs offline workflow induction, distilling messy, execution-level logs into \emph{reusable and actionable procedures}. Unlike prior work that summarizes successful traces into narrative guidance or local rules, WISE-Flow induces a \emph{structured workflow with explicit, environment-grounded prerequisites}, leveraging \emph{contrastive} evidence from clean successes, error-recovered successes, and failures. Even with complete service logs, raw trajectories are often not directly reusable as procedures. They contain detours and errors, and different trajectories may succeed via different paths. The key challenge is to recover a \emph{stable} procedure with \emph{executable constraints} such as prerequisites and ordering. WISE-Flow turns outcome-divergent traces into supervision, where contrastive comparisons isolate the minimal deviations that separate success from failure and encode them as explicit ordering constraints and prerequisites in the workflow.

\noindent\textbf{Pre-processing and contrast construction.}
We partition trajectories for each task into three categories: \emph{clean successes} (no tool-call errors), \emph{recovered successes} (contain tool-call errors but later recover), and \emph{failures}. We then construct paired contrasts by pairing a clean success with a recovered success or a failure. Concatenating each pair forms a contrastive block $\mathcal{C}(\mathcal{T})$, which makes the outcome-determining differences explicit. The block highlights \textbf{effective action ordering} in successes and reveals violated \textbf{prerequisites and ordering constraints} via environment feedback in the recovered or failed traces. Meanwhile, we use an LLM to produce a task summary $s(\mathcal{T})$ from the user queries, as shown in the Appendix Figure~\ref{fig:prompt-wiseflow-sum-precheck}.

\noindent\textbf{Workflow induction.}
WISE-Flow induces a workflow via $w = g(s(\mathcal{T}), \mathcal{C}(\mathcal{T}))$,
where $g(\cdot)$ is an LLM-based workflow inducer that takes the task summary $s(\mathcal{T})$ and the contrastive block $\mathcal{C}(\mathcal{T})$ as input, and outputs a workflow $w$ under a constrained JSON schema (Appendix Fig.~\ref{fig:workflow_induction_d}). Intuitively, $s(\mathcal{T})$ describes the global task intent, while $\mathcal{C}(\mathcal{T})$ provides discriminative supervision about what mattered for success, which is inspired by AutoGuide~\cite{fu2024autoguide} but targets a richer procedural representation than local decision rules. Directly generating procedures from logs is brittle, as LLMs may hallucinate steps, omit execution-critical constraints, or make ungrounded recommendations. Therefore, we propose the \textbf{grounded induction via multi-pass verification}, which separates evidence extraction from synthesis and verifies support in the input trajectories. Specifically, we use a \textbf{three-pass procedure} (Appendix Fig.~\ref{fig:workflow_induction}): (i) \textbf{analysis} identifies the goal-consistent action sequence and the key environment feedback explaining failures; (ii) \textbf{drafting} synthesizes a candidate workflow with explicit steps and per-action prerequisites; and (iii) \textbf{reflection and revision} validates each step, prerequisite, and branching condition against trajectories, removing or fixing unsupported or non-executable recommendations. This verification-based design improves faithfulness and executability over single-pass generation.

\noindent\textbf{Workflow representation.}
To bridge the retrieval-to-action gap, we design a workflow representation that couples a \emph{task-level workflow} with \emph{prerequisite-augmented action blocks} (see Appendix Fig.~\ref{fig:workflow_induction_d} and Fig.~\ref{fig:workflow_sample}). The workflow backbone summarizes the task with a short description and ordered milestones, which support progress alignment. Each action block specifies an action with explicit prerequisites (both global and scenario-specific) and conditional next-step transitions grounded in environment feedback. Prerequisites act as execution guards against common procedural mistakes, and transitions encode branching behavior observed in past trajectories. As a result, retrieved experience becomes actionable guidance with explicit success/recovery routes and checkable prerequisites, enabling feasibility-aware next-step selection.

\subsection{Online Workflow-guided Agent Control}
In the third stage, WISE-Flow assists conversational agents during deployment with the induced workflow library  (see the prompt in Appendix Figure~\ref{fig:prompt-wiseflow-tau2}). Retrieving past experience is not enough. The challenge is that the agent must (i) localize its current stage within a multi-step procedure and (ii) select a next action that is feasible under the current world state. Without explicit progress alignment and prerequisite checking, retrieved experience provides limited support for a feasible next action. WISE-Flow therefore converts retrieved workflows into actionable next-step guidance via \textbf{progress alignment and prerequisite-aware feasibility checking}. Let $h_t$ denote the current interaction history up to turn $t$: $h_t = (e_1, e_2, ..., e_t)$. Then, we extract from $h_t$ a summary of the user's request as the task-level query $q_t$ via an LLM summarizer, and identify the current tool trace as well as the last successfully executed tool $a_{t-1}$.

\noindent\textbf{Workflow retrieval.} Each induced workflow $w$ is a structured object, which contains the task description, entry steps, planned steps as well as action blocks, and it is indexed with an embedding $\mathbf{v}_w$. We embed the task query as $\mathbf{u}_t=\phi(q_t)$ and use it to retrieve top-$K$ workflows as follows: $\mathcal{W}_t = TopK_{w\in\mathcal{W}} <norm(\mathbf{u}_t), norm(\mathbf{v}_w)>$, where $\phi(\cdot)$ is the embedding model and $\mathcal{W}$ is the workflow library.

\noindent\textbf{Progress alignment.} Retrieval provides relevant workflows but does not specify where the agent is within a procedure. Therefore, we localize progress by aligning the last successful tool call $a_{t-1}$ to the planned-step list $S^{(w)}$. Specifically, we select the most similar step in $S^{(w)}$ and take its subsequent step as the next-step candidate. This converts retrieved workflows into stage-specific suggestions and reduces the next-action space to a small action set that is both task-relevant and procedurally coherent.

\noindent\textbf{Prerequisite-aware action validation.} Stage-conditioned suggestions may be infeasible due to unmet prerequisites. Using prerequisite annotations in retrieved action blocks, we check each candidate action against the interaction trace and environment feedback observed so far (see the prompt in Appendix Figure~\ref{fig:prompt-wiseflow-sum-precheck}). Then, we return candidates with prerequisite status (met or unmet), guiding the agent to either execute the action directly or resolve missing prerequisites first. This makes feasibility explicit, instead of requiring the model to infer constraints from scratch.

\noindent\textbf{Workflow-conditioned context injection.} Finally, we construct a compact structured context block, which contains the retrieved workflows, suggested next action candidates, and prerequisite status for each candidate. Compared with conventional context injection, it provides actionable next-step guidance with checkable constraints, enabling more reliable tool execution under the current world state.

\section{Experimental Setup}
We evaluate the performance of WISE-Flow through several controlled experiments, focusing on: 
\textbf{(RQ1)} End-to-end performance gains of WISE-Flow.
\textbf{(RQ2)} The impact of modeling complete service logs.
\textbf{(RQ3)} The effect of structured experience representation.
\textbf{(RQ4)} The benefit of aggregating trajectories per task for workflow induction.

\subsection{Benchmarks and Environments}
We use two controllable simulation environments to collect interaction trajectories as past experience for offline workflow induction, and we evaluate our method in the same environments.

\noindent\textbf{ToolSandbox}~\cite{lu2025toolsandbox} provides an executable, stateful service environment for multi-turn tool use, where tool invocations induce persistent state transitions and produce fine-grained feedback. Intermediate environment states are observable. An agent's success in ToolSandbox depends on executing the correct tool calls in the right order. These properties make it well-suited for our setting. Our evaluation is done on 76 tasks of ToolSandbox spanning both multiple-tool-call scenarios and multiple-user-turn scenarios.

\noindent\textbf{$\tau^2$-bench}~\cite{barres2025tau} evaluates conversational agents in a dual-control setting, where both the agent and the user can invoke tools to update a shared world state. It targets collaborative troubleshooting that demands coordination and grounded communication, while exposing procedural failures and recovery cues via environment-side feedback. In our experiments, we evaluate agents on the telecom domain, the most complex domain of the $\tau^2$-bench, which contains 114 tasks.

\subsection{Baselines}
We compare WISE-Flow with five representative strategies for agent design (see Appendix~\ref{app:exp_setup_baseline}). 

\noindent\textbf{LLM-only} agent generates responses solely based on the current dialog context, without any explicit tool-use control loop, reflection mechanism, or external memory. It relies entirely on the backbone LLM’s implicit reasoning and instruction-following capabilities.

\noindent\textbf{ReAct}~\cite{yao2022react} structures agent decision making as an interleaved reasoning–acting loop. The agent alternates between natural language reasoning and tool invocations, using tool observations to revise its intermediate plans. ReAct improves within-episode control through immediate feedback, but does not provide an explicit mechanism for cross-episode experience reuse. In our implementation, we prompt the agent to follow the standard Thought-Action-Observation format.

\noindent\textbf{Reflexion}~\cite{shinn2023reflexion} equips an agent with post-hoc self-reflection. After each episode, an LLM analyzes the trajectory, identifies failure modes, and summarizes lessons as textual memories. For a new task, the agent retrieves relevant reflections and conditions its reasoning on them, aiming to avoid repeating past mistakes and improve cross-episode performance. In our implementation, we follow this protocol by generating reflection summaries from historical service logs and injecting the retrieved reflections into the agent context at test time.

\noindent\textbf{REMEMBERER}~\cite{zhang2023large} constructs an experience memory from past trajectories. REMEMBERER maintains a persistent tabular memory, where each record contains a task goal, an observation, an action taken under that observation, and a value estimate (e.g., a $Q$-value) for that action. At inference time, given the current task and observation, the agent retrieves the most similar records and uses their value estimates to rank and bias action selection, preferring high-value actions while discouraging lower-value alternatives among the retrieved candidates.

\noindent\textbf{AutoGuide}~\cite{fu2024autoguide} distills offline agent trajectories into concise, context-aware guidelines expressed as conditional rules, where the context is the summary of the current status. At test time, the agent retrieves and follows these guidelines to bias action choices under similar contexts.

\subsection{Models}
We use Claude 3.7 Sonnet~\cite{anthropic_claude37_systemcard_2025} and Qwen3-235B-A22B~\cite{qwen3technicalreport} as the base models for our agents, and use Claude 3.7 Sonnet for user simulation. We set the temperature to $0.9$ when collecting diverse experience, and to $0$ during testing. More details are in Appendix~\ref{app:exp_setup}.
 
\subsection{Evaluation Metric}
We evaluate the agent on both benchmarks using three metrics: (1) the average task success rate, (2) the average missed-milestone ratio, and (3) the $F_\beta$ score. Let $N$ be the number of task. The average success rate is $SR=\frac{1}{N} \sum_{i=1}^{N}I_{[\textit{task}_i\textit{ is solved}]}$, where $I$ is the indicator function. The average missed-milestone ratio is 
    $MMR=\frac{1}{N} \sum_{i=1}^{N}\frac{m_i^{miss}}{m_i}$,
where $m_i$ denotes the required milestones for task $i$ and $m_i^{miss}$ denotes those not achieved.

Most evaluations of conversational agents emphasize end-to-end success~\cite{barres2025tau, lu2025toolsandbox}, such as the task success rate, but often ignore action errors during execution. In user-facing services, reliability (i.e., steady progress without failures) matters as much as the final outcome. Therefore, we introduce $F_{\beta}$~\cite{Chinchor1992MUC4EM} into the evaluation of tool-augmented conversational agents to to jointly capture milestone coverage and execution reliability. Similar to $F_1$, $F_{\beta}$ is a weighted harmonic mean of milestone recall $R$ and an error-aware precision term $P$:
\begin{equation}
    F_{\beta}=\frac{(1+\beta^2)\,P*R}{\beta^2 P+R},
\end{equation}

where $\beta$ is a metric hyperparameter that adjusts the relative importance of recall versus precision. $\beta>1$ emphasizes recall. In the following results, we report $F_\beta$ with $\beta=5$, since failing to cover required milestones is more harmful than making occasional errors. We define
\begin{align}
    R = \frac{m_i - m_i^{miss}}{m_i}, \ P = \frac{|A_i|-|E_i|}{|A_i|}.
\end{align}
where $|A_i|$ denotes the total number of actions made by the agent in task $i$, and $|E_i|$ denotes the number of action errors encountered in task $i$. 

To better understand how the agent succeeds, we further break down successful trials into \textbf{one-shot successes} (solved without trial-and-error) versus \textbf{successes after iterative correction}. We quantify the reliance on trial-and-error by computing the proportion of trial-and-error successes among all successful trials: $\mathrm{TE\text{-}ratio} = \frac{\text{\# (success after trial-and-error)}}{\text{\# (all successful trials)}}$.

Besides these metrics, we also report the default metrics provided by each benchmark, i.e., the similarity score of ToolSandbox, which measures the similarity between the agent’s actions and resulting world states and the ground truth, and $pass\textasciicircum{}k$ of $\tau^2$-bench~\cite{yao2024tau}, which defines the probability that an agent succeeds on the same task for $k$ i.i.d. trials (i.e., all $k$ runs are successful), averaged across tasks.

\section{Results and Analysis}
\subsection{Overall Results (RQ1)}
\label{sec: overall-dyn}
\begin{table*}[th!]
\centering
\begin{tabular}{|c|l|c|c|c|c|}
\cline{1-6}
Model                                               & \multicolumn{1}{c|}{Method}     & \multicolumn{1}{l|}{Similarity $\uparrow$} & \multicolumn{1}{c|}{$F_\beta\uparrow$} & \multicolumn{1}{l|}{MMR $\downarrow$} & \multicolumn{1}{l|}{SR $\uparrow$}  \\ \cline{1-6}
\multirow{6}{*}{Qwen3-235B-A22B}                              & LLM-only   & 0.856                           & 0.8985                       & 9.75\%                                        & 77\%                                \\ \cline{2-6}
                                                    & ReAct      & 0.858                           & 0.9008                       & 9.70\%                                        & 76\%                               \\ \cline{2-6}
                                                    & Reflexion  & 0.866                           & 0.8999                       & 9.79\%                                        & 84\%                              \\ \cline{2-6}
                                                    & REMEMBERER & 0.871                           & 0.9134                       & 8.51\%                                        & 78\%                               \\ \cline{2-6}
                                                    & AutoGuide  &  0.911*                           &  0.9484*                       &  5.08\%*                                       &  87\%*                               \\ \cline{2-6}
                                                    & WISE-Flow  & $ {\mathbf{0.923}}$                           & $ {\mathbf{0.9615}}$                       & $ {\mathbf{3.69\%}}$                                        & $ {\mathbf{88\%}}$                               \\ \cline{1-6}
\multicolumn{1}{|c|}{\multirow{6}{*}{Claude 3.7 Sonnet}} & LLM-only   & 0.871                           & 0.9277                       & 6.99\%                                        & 87\%                                \\ \cline{2-6}
\multicolumn{1}{|c|}{}                              & ReAct      &  0.911*                           & 0.9603                       & 3.72\%                                        & 92\%                                \\ \cline{2-6}
\multicolumn{1}{|c|}{}                              & Reflexion  & 0.875                           & 0.9462                       & 5.42\%                                        & 87\%                                \\ \cline{2-6}
\multicolumn{1}{|c|}{}                              & REMEMBERER & $ {\mathbf{0.914}}$                          &  0.9728*                       &  2.47\%*                                        & $ {\mathbf{95\%}}$                                \\ \cline{2-6}
\multicolumn{1}{|c|}{}                              & AutoGuide  &  0.911*                           & 0.9699                       & 2.90\%                                        & 91\%                                \\ \cline{2-6}
\multicolumn{1}{|c|}{}                              & WISE-Flow  & 0.910                           & $ {\mathbf{0.9745}}$                      & $ {\mathbf{2.46\%}}$                                        &  95\%*                                \\ \cline{1-6}
\end{tabular}
\caption{Results on ToolSandbox. Best results are in bold and second-best are with * . Same for all tables. Standard deviations are reported in Appendix Table~\ref{tab:all_toolsandbox_std}.}
\label{tab:all_toolsandbox}
\end{table*}

% \usepackage{multirow}
\begin{table*}[ht!]
\centering
\begin{tabular}{|c|l|c|c|c|c|c|}
\hline
Model                                               & \multicolumn{1}{c|}{Method}     & pass\textasciicircum{}1 $\uparrow$ & pass\textasciicircum{}2 $\uparrow$ & pass\textasciicircum{}3 $\uparrow$ & $F_\beta \uparrow$ & MMR $\downarrow$ \\ \hline
\multirow{6}{*}{Qwen3-235B-A22B}                              & LLM-only   &$ {\mathbf{0.395}}$                   &  0.222*                   & 0.132                   &  0.5537*  &  43.84\%*                  \\ \cline{2-7} 
                                                    & ReAct      & 0.348                   &  0.222*                   &  0.184*                   & 0.4217  & 57.50\%                  \\ \cline{2-7} 
                                                    & Reflexion  & 0.345                   & 0.216                   & 0.167                   & 0.5456  & 45.66\%                  \\ \cline{2-7} 
                                                    & REMEMBERER & 0.319                   & 0.164                   & 0.097                   & 0.4792  & 52.05\%                  \\ \cline{2-7} 
                                                    & AutoGuide  & 0.322                   & 0.211                   & 0.149                   & 0.4907  & 51.12\%                  \\ \cline{2-7} 
                                                    & WISE-Flow  & $ {\mathbf{0.395}}$                    & $ {\mathbf{0.252}}$                    & $ {\mathbf{0.193}}$                   & $ {\mathbf{0.6343}}$  & $ {\mathbf{36.65\%}}$                  \\ \hline
\multicolumn{1}{|c|}{\multirow{6}{*}{Claude 3.7 Sonnet}} & LLM-only   & 0.462                   & 0.339                   & 0.272                   & 0.6329  & 36.92\%                  \\ \cline{2-7} 
\multicolumn{1}{|c|}{}                              & ReAct      & 0.506                   & 0.359                   & 0.278                   & 0.6357  & 36.59\%                  \\ \cline{2-7} 
\multicolumn{1}{|c|}{}                              & Reflexion  &  0.535*                   &  0.433*                   &  0.377*                   & 0.6540  & 34.65\%                  \\ \cline{2-7} 
\multicolumn{1}{|c|}{}                              & REMEMBERER & 0.523                   & 0.424                   & 0.368                   &  0.6629*  &  33.92\%*                  \\ \cline{2-7} 
\multicolumn{1}{|c|}{}                              & AutoGuide  & 0.453                   & 0.366                   & 0.333                   & 0.6067  & 39.62\%                  \\ \cline{2-7} 
\multicolumn{1}{|c|}{}                              & WISE-Flow  & $ {\mathbf{0.564}}$                   & $ {\mathbf{0.468}}$                   & $ {\mathbf{0.421}}$                   & $ {\mathbf{0.6936}}$  & $ {\mathbf{30.75\%}}$                  \\ \hline
\end{tabular}

\caption{Results on $\tau^2$-bench. Standard deviations are reported in Appendix Table~\ref{tab:all_tau2_std}.}
\vspace{-11pt}
\label{tab:all_tau2}
\end{table*}

We simulate 10 independent trials to collect the raw experience, i.e., the initial trajectories, for each task. During evaluation, the agent retrieves the top $K$ most similar experiences to guide its actions. We set $K=3$, run each task 10 times on ToolSandbox and 3 times on the $\tau^2$-bench, and report the performance averaged over all tasks and trials. The overall results are shown in Table~\ref{tab:all_toolsandbox} and Table~\ref{tab:all_tau2}. For $\tau^2$-bench, since the $pass\textasciicircum{1}$ is equivalent to the success rate, we omit the success rate in Table~\ref{tab:all_tau2}.

\begin{figure}
    \centering
    \includegraphics[width=1\linewidth]{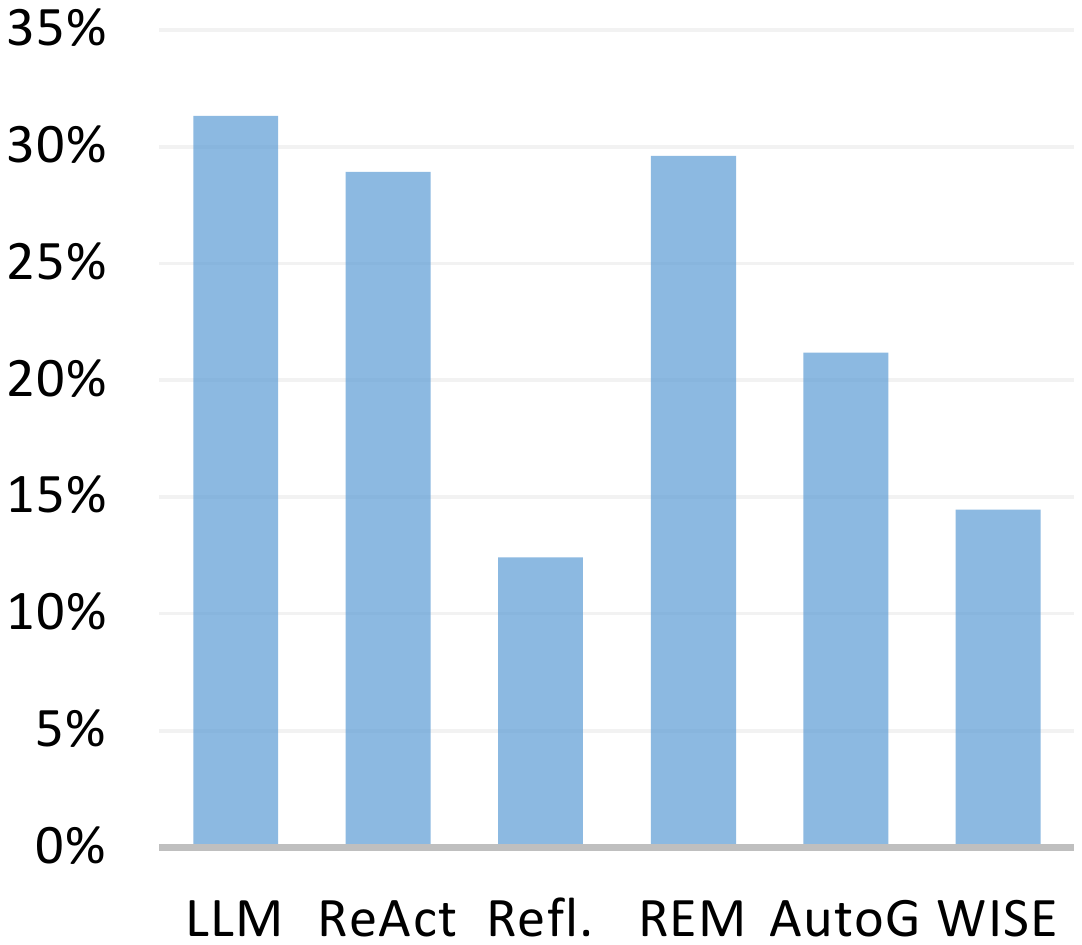}
    \vspace{-33pt}
    \caption{Proportion of trial-and-error (TE) successes among all successful trials (lower is better). We decompose successful episodes into one-shot successes and successes after TE, and report the proportion of the latter for each method.(Refl. = Reflexion, REM = REMEMBERER, AutoG = AutoGuide, WISE= WISE-Flow)}
    \label{fig:toolsandbox_recover_claude}
    \vspace{-11pt}
\end{figure}

On ToolSandbox, WISE-Flow achieves the best overall performance when using Qwen3-235B-A22B as the base model, obtaining the highest success rate and the lowest missed-milestone ratio. While its performance with Claude 3.7 Sonnet is also competitive, its missed-milestone ratio is only $0.01\%$ higher than the best-performing baseline, REMEMBERER. We then compare the $TE-ratio$ of all methods in Figure~\ref{fig:toolsandbox_recover_claude}. We find that WISE-Flow's one-shot success rate is substantially higher than REMEMBERER's, suggesting that our method not only improves final task completion but also reduces error-prone exploration during execution. We also observe that Reflexion has a relatively low proportion of successes achieved after trial-and-error. This may be because its memory stores explicit reflections on failures, which are easier for the agent to interpret and can help it avoid repeating the same mistakes. However, Reflexion lacks the global planning mechanism used in WISE-Flow, which limits its ability to solve complex tasks end-to-end and results in a lower overall success rate. Overall, WISE-Flow achieves the best balance between high success and low error rate, which is also consistently reflected by its $F_\beta$ score. That also shows $F_\beta$ provides a more holistic and discriminative evaluation of agent performance than any single metric alone.

On $\tau^2$-bench, WISE-Flow achieves the best performance in terms of $pass\textasciicircum{k}$ for $k=1,2,3$, $F_\beta$, and missed-milestone ratio, when using Claude 3.7 Sonnet as the base model. When using Qwen3-235B-A22B, all other baselines are even worse than the LLM-only method at $pass\textasciicircum{1}$, but they are generally more robust. As $k$ increases, their $pass\textasciicircum{k}$ degrades more slowly and can surpass the LLM-only baseline. In contrast, WISE-Flow consistently achieves the highest $pass\textasciicircum{k}$ for all $k\in\{1,2,3\}$ and attains a substantially higher $F_\beta$ than the other methods, indicating a better balance between milestone coverage and error control.

Overall, WISE-Flow achieves consistently strong performance across different backbones and environments.

\subsection{Effectiveness of Backend Signals (RQ2)}
Next, we investigate the effect of using full logs as past experience, which capture not only user-agent interactions but also the agent’s actions and environment feedback, such as error messages. 

We use WISE-Flow to extract two different workflows: one is extracted from full logs, and the other is extracted from the logs without environment feedback. We then apply these two workflow variants to the agents and compare their performance. We use ground-truth-based retrieval, meaning that the retrieved workflow comes from the same task instance, thereby eliminating retrieval randomness. We conduct this experiment on ToolSandbox, running each task for 10 trials. 

From Table~\ref{tab:eval_envfb}, we can observe a clear performance gap between the two variants. Applying workflows extracted from full logs achieves more milestones than those extracted without environment feedback, improving milestone coverage by around $3\%$. It also achieves a higher $F_\beta$ score by $3\%$, which indicates better overall performance that jointly accounts for milestone completion and error control. We further observe that the overall success rate increases by $3.9\%$, while the proportion of trials that succeed only after trial-and-error decreases by $3\%$, suggesting that the improvement is not only in final task completion, but also in one-shot success. This further demonstrates that the gain in $F_\beta$ is driven not merely by a higher success rate, but also by reduced errors during execution.

Overall, these results demonstrate that environment feedback is essential when using past experience for agent guidance. The trial-and-error signals embedded in tool outputs and error messages provide valuable supervision, enabling agents to complete tasks more effectively while requiring fewer corrective attempts.

\begin{table}[h!] \small %\color[HTML]{ED7D31}
\centering
\begin{tabular}{|l|c|c|c|}
\hline
        & \multicolumn{1}{c|}{\makecell[c]{Full\\Trace}} & \multicolumn{1}{c|}{\makecell[c]{Dialogue-Only\\Trace}} & $\Delta$ \\ \hline
Similarity $\uparrow$               & $ {\mathbf{0.9136}}$   & 0.8852                                   &  $\downarrow$ 2.8\%                 \\ \hline
$F_\beta$ $\uparrow$  & $ {\mathbf{0.9755}}$   & 0.9457                                   & $\downarrow$ 3.0\%                 \\ \hline
MMR $\downarrow$ & $ {\mathbf{2.5\%}}$    & 5.4\%                                    & $\uparrow$ 3.0\%                 \\ \hline
SR $\uparrow$             & $ {\mathbf{93.5\%}}$                          & 89.6\%                                   & $\downarrow$ 3.9\%                 \\ \hline
\makecell[l]{SR with \\ trial-and-error $\downarrow$}           & $ {\mathbf{8.5\%}}$                          & 11.5\%                                   & $\uparrow$ 3.0\%                 \\ \hline
\end{tabular}
\caption{Ablation on log completeness: workflows extracted from full traces vs. dialogue-only traces. $\Delta$ denotes the per-row difference between the two settings.}
\vspace{-11pt}
\label{tab:eval_envfb}
\end{table}

\begin{table*}[th!]
\centering
\begin{tabular}{|l|c|c|c|c|c|}
\hline
Method                                   & \multicolumn{1}{l|}{Similarity $\uparrow$} & \multicolumn{1}{c|}{$F_\beta$ $\uparrow$} & \multicolumn{1}{c|}{MMR $\downarrow$} & \multicolumn{1}{c|}{SR $\uparrow$} & \multicolumn{1}{c|}{SR in one-shot $\uparrow$} \\ \hline
LLM-only                                  & 0.8707                          & 0.9277                       & 6.99\%                   & 86.97\%                 & 59.74\%                             \\ \hline
Raw  Logs                        & 0.8879                          & 0.9449                       & 5.45\%                   & 88.88\%                 & 71.63\%                             \\ \hline
Plain-text   Workflow    & 0.8989                          & 0.9584                       & 4.06\%                   & 93.00\%                 & 70.88\%                           \\ \hline
WISE-Flow                 & $ {\mathbf{0.9136}}$                          & $ {\mathbf{0.9755}}$                       & $ {\mathbf{2.46\%}}$                   & $ {\mathbf{93.50\%}}$                 & $ {\mathbf{85.00\%}}$                             \\ \hline
\end{tabular}
\caption{Effect of experience representation on ToolSandbox.}
\vspace{-11pt}
\label{tab:rep_wf}
\end{table*}
\subsection{Benefits of Structured Experience Representations (RQ3)}
In this section, we investigate the impact of structuring past experience for effective agent guidance. We compare three forms of experience: (i) raw service logs, (ii) a plain-text workflow description that captures the order of actions in the workflow, and (iii) our structured WISE-Flow representation. We conduct this experiment on ToolSandbox with Claude 3.7 Sonnet as the base model, and use ground-truth-based retrieval, i.e., the retrieved experience comes from the same task, thereby eliminating retrieval randomness and isolating the effect of representation.

From Table~\ref{tab:rep_wf}, we observe that experience representation plays a critical role in downstream performance. Both Raw Logs and Plain-text Workflow outperform the LLM-only baseline across all metrics, indicating that leveraging past experience is beneficial. Among them, WISE-Flow achieves the best results overall, with the highest similarity, success rate, and $F_\beta$, as well as the lowest missed-milestone ratio. Notably, while the plain-text workflow description improves final task success, it does not improve one-shot success over raw conversations, suggesting that unstructured or loosely structured experience may still lead to trial-and-error execution. In contrast, WISE-Flow substantially increases one-shot success, demonstrating that structured experience not only helps agents complete tasks but also reduces corrective attempts by better guiding action selection and error avoidance.

\subsection{Benefits of Aggregating Trajectories for Workflow Construction (RQ4)}
\label{sec:exp_workflow_aggregating}
We then show that aggregating multiple trajectories from the same task leads to better workflows for agent guidance. Given several trajectories collected for a task, we compare two strategies for the workflow induction of WISE-Flow: (i) \textbf{Trajectory-wise} workflow extraction, which constructs one workflow from each individual trajectory, and (ii) \textbf{Task-wise} workflow extraction, which aggregates trajectories from the same task and extracts a single consolidated workflow. We evaluate the downstream performance of agents on ToolSandbox using workflows produced by these two strategies under the same retrieval method and evaluation setup.

\begin{figure}[h!]
    \centering
    \includegraphics[width=1\linewidth]{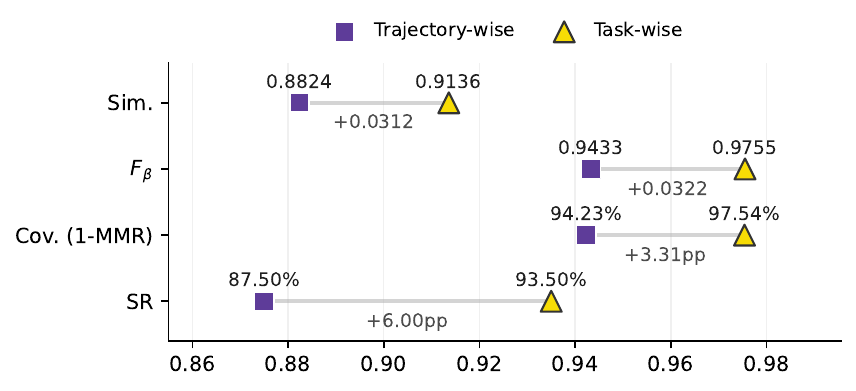}
    \vspace{-22pt}
    \caption{Trajectory-wise vs. task-wise workflow induction on ToolSandbox. Higher Value is better. (We report milestone coverage as $1-\mathrm{MMR}$. Sim. = Similarity.)}
    \label{fig:traj-vs-task-claude37}
\end{figure}

The results show a clear advantage of task-wise workflow induction over trajectory-wise induction. As shown in Figure~\ref{fig:traj-vs-task-claude37}, aggregating multiple trajectories from the same task yields consistent improvements across all metrics, leading to about a $3\%$ gain in similarity, $F_\beta$, and milestone coverage, as well as a $6\%$ gain in success rate. We believe this improvement mainly comes from aggregating and comparing multiple trajectories during workflow induction. A workflow induced from a single trajectory can include some unnecessary or even false steps in the trial, whereas task-wise induction lets the LLM compare different attempts and summarize a more stable, task-relevant action pattern into a consolidated workflow. This is more effective than presenting all trajectories to the agent and asking it to perform the comparison during task execution. As a result, the agent receives clearer guidance, makes fewer mistakes, and achieves higher performance.

Besides the observations above, our supplementary analysis also demonstrates that:
(i) the proposed three-pass procedure produces stronger induced workflows compared to single-shot induction (Appendix~\ref{sec:exp_multi-pass});
(ii) WISE-Flow workflows generalize across tasks, enabling performance gains when retrieving workflows from other tasks (Appendix~\ref{sec:exp_cross_task});
and (iii) WISE-Flow retrieves genuinely useful workflows and benefits from them (Appendix~\ref{sec:exp_retrieval_effect}).

\section{Conclusion}
In this paper, we introduce WISE-Flow, a framework that turns raw service logs into structured, reusable workflows and guides tool-augmented conversational agents via retrieval at both the workflow and action levels. We also propose a new workflow representation and introduce a new evaluation metric $F_\beta$. Experiments on ToolSandbox and $\tau^2$-bench show that WISE-Flow improves performance and better balances success and errors, as reflected by higher $F_\beta$. Our analyses indicate the importance of environment feedback and show that structured, trajectory-aggregated experience provides stronger guidance and can transfer across tasks, even without the same-task experience. Overall, WISE-Flow makes past tool-use experience more usable and effective for LLM-based agents in conversational service environments.

% \clearpage % It says "without page breaks" https://acl-org.github.io/ACLPUB/formatting.html#limitations
\section*{Limitations}
Despite its effectiveness, WISE-Flow has several limitations.

\noindent\textbf{Dependence on log observability.} WISE-Flow assumes access to complete service trajectories that include not only dialog turns but also tool traces and environment-side feedback. In many real deployment, such information may be partially missing. When the environment-side feedback is incomplete, the induced workflow, along with action prerequisites and procedural constraints, becomes harder to recover and may not fully capture the successful strategies, which may reduce the benefit of workflow guidance.

\noindent\textbf{Reliability of offline workflow induction.} The workflow induction of WISE-Flow relies on an LLM to synthesize structured procedures from trajectories. Although we employ a multi-pass induction procedure to reduce unsupported steps, the inducer may still introduce spurious ordering constraints, include unnecessary steps, or hallucinate prerequisites that appear plausible but are not strictly necessary, which can be harmful in online deployment, especially if the agent treats them as strong guidance. Incorporating executable verification or learning minimal sufficient procedures remains an important direction.

\noindent\textbf{Progress localization and branching complexity.} The online control of WISE-Flow localizes agent progress by aligning recent successful tool calls to the planned steps of retrieved workflows and proposing stage-specific next actions. This heuristic can be brittle in workflows with substantial branching, repeated actions across stages, or non-sequential interleavings of tool calls. In such cases, matching only a short tool trace may misidentify the current stage and narrow the action shortlist incorrectly. Extending progress alignment to sequence-level matching, state-aware alignment, or probabilistic stage tracking may improve robustness for more complex tasks.

\section*{Ethical Considerations}
Our research complies with applicable ethical standards and relies exclusively on publicly available datasets, benchmarks, and environments. All simulations are performed strictly within the scope, scenarios, and constraints specified by these resources. The contents of the datasets, benchmarks, and environments are provided by their respective sources and do not reflect the views, opinions, or positions of the authors.

\paragraph{Potential Risks and Mitigations.}
Our work is evaluated using ToolSandbox and $\tau^2$-bench, which are controlled environments and benchmarks. While these resources are intended for safe and reproducible research, the proposed WISE-Flow framework can introduce dual-use and unintended harmful effects if deployed outside controlled settings. First, the ability to induce and guide multi-step tool usage workflows could be repurposed to automate undesirable or abusive actions, e.g., large-scale API abuse, unauthorized automation of workflows in sensitive systems, even if the original evaluation is in simulation. Second, errors in workflow induction or progress localization may lead an agent to issue inappropriate tool calls. In real environments connected to external services, such guidance errors could cause irreversible effects, waste resources, or trigger unsafe behaviors. 

We mitigate these potential harms by (i) emphasizing that our evaluation is limited to simulated sandbox environments and not intended for direct deployment in high-stakes real-world systems, (ii) recommending restrictions and confirmation steps for high-impact or irreversible tools, (iii) sanitizing or abstracting sensitive fields in task traces used for workflow induction, and (iv) encouraging future work to incorporate executable verification, safety filters, and fairness evaluations across diverse domains. These safeguards help ensure that workflow guidance enhances benign automation without exacerbating misuse, privacy violations, or unfair outcomes across stakeholder groups.

% This document does not cover the content requirements for ACL or any
% other specific venue.  Check the author instructions for
% information on
% maximum page lengths, the required ``Limitations'' section,
% and so on.

% \section*{Acknowledgments}


\bibliography{custom}

\appendix

\section{Appendix}
\label{sec:appendix}

\subsection{Experimental Setup}
\label{app:exp_setup}
\subsubsection{Historical Experience Simulation}
When simulating the historical experiences, we set the temperature as $0.9$. For user simulation, we use Claude 3.7 Sonnet~\cite{anthropic_claude37_systemcard_2025}. For agent simulation, we Claude 3.7 Sonnet ("us.anthropic.claude-3-7-sonnet-20250219-v1:0") and Qwen3-235B-A22B~\cite{qwen3technicalreport} ("qwen.qwen3-235b-a22b-2507-v1:0") as the base models. We run 10 trials per benchmark per agent using the environment's default prompts.

\subsubsection{Implementation of Baselines}
\label{app:exp_setup_baseline}
\begin{figure*}[ht!]
    \centering
    \includegraphics[width=1\linewidth]{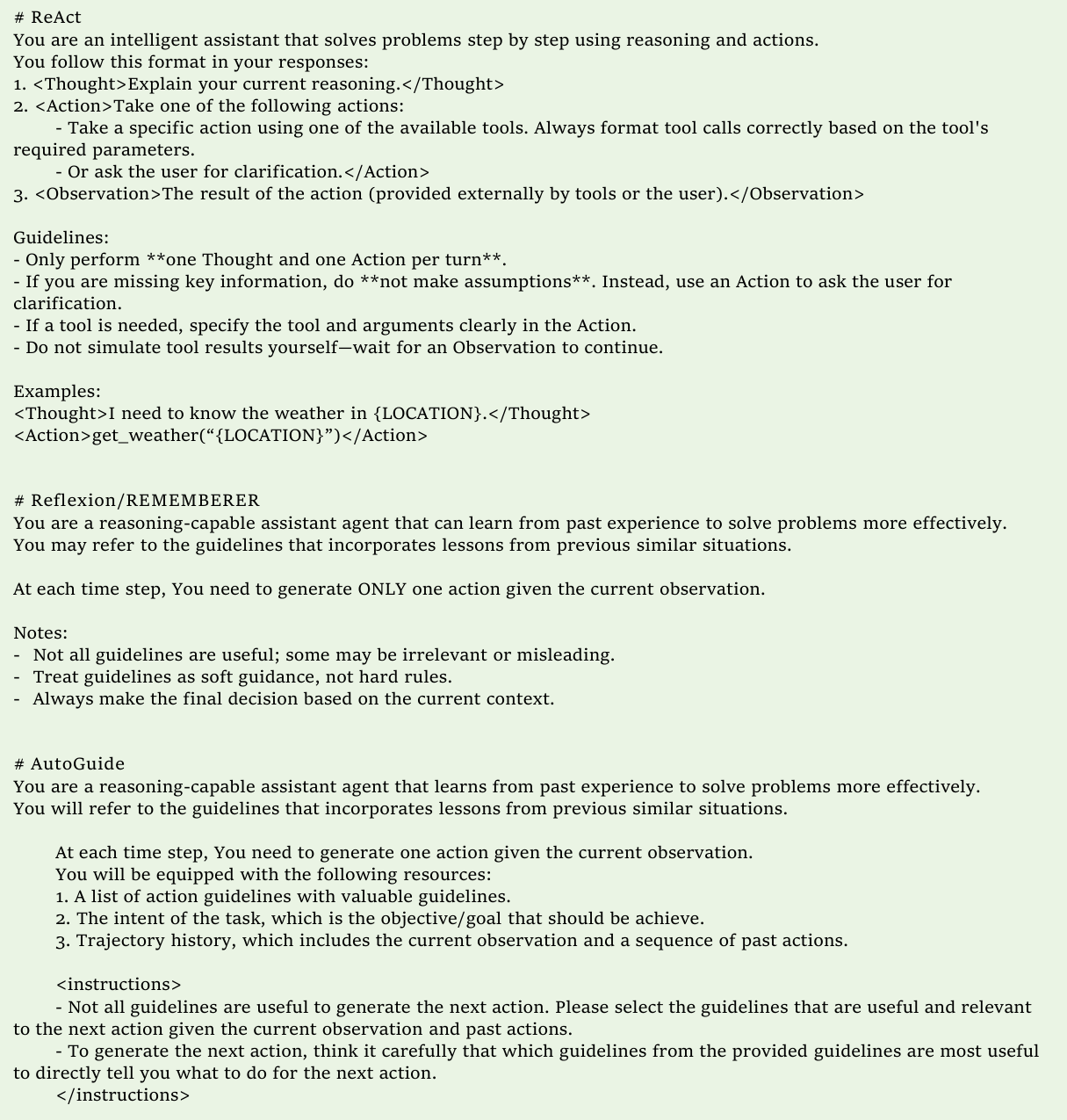}
    \caption{Prompts of all baselines on ToolSandbox.}
    \label{fig:prompt_baseline_toolsandbox}
\end{figure*}

The prompts of the baselines in ToolSandbox are shown in Figure~\ref{fig:prompt_baseline_toolsandbox}.

\subsubsection{Implementation of WISE-Flow}
\begin{figure*}[t]
  \centering

  \begin{subfigure}{0.49\linewidth}
    \centering
    \includegraphics[page=1,width=\linewidth]{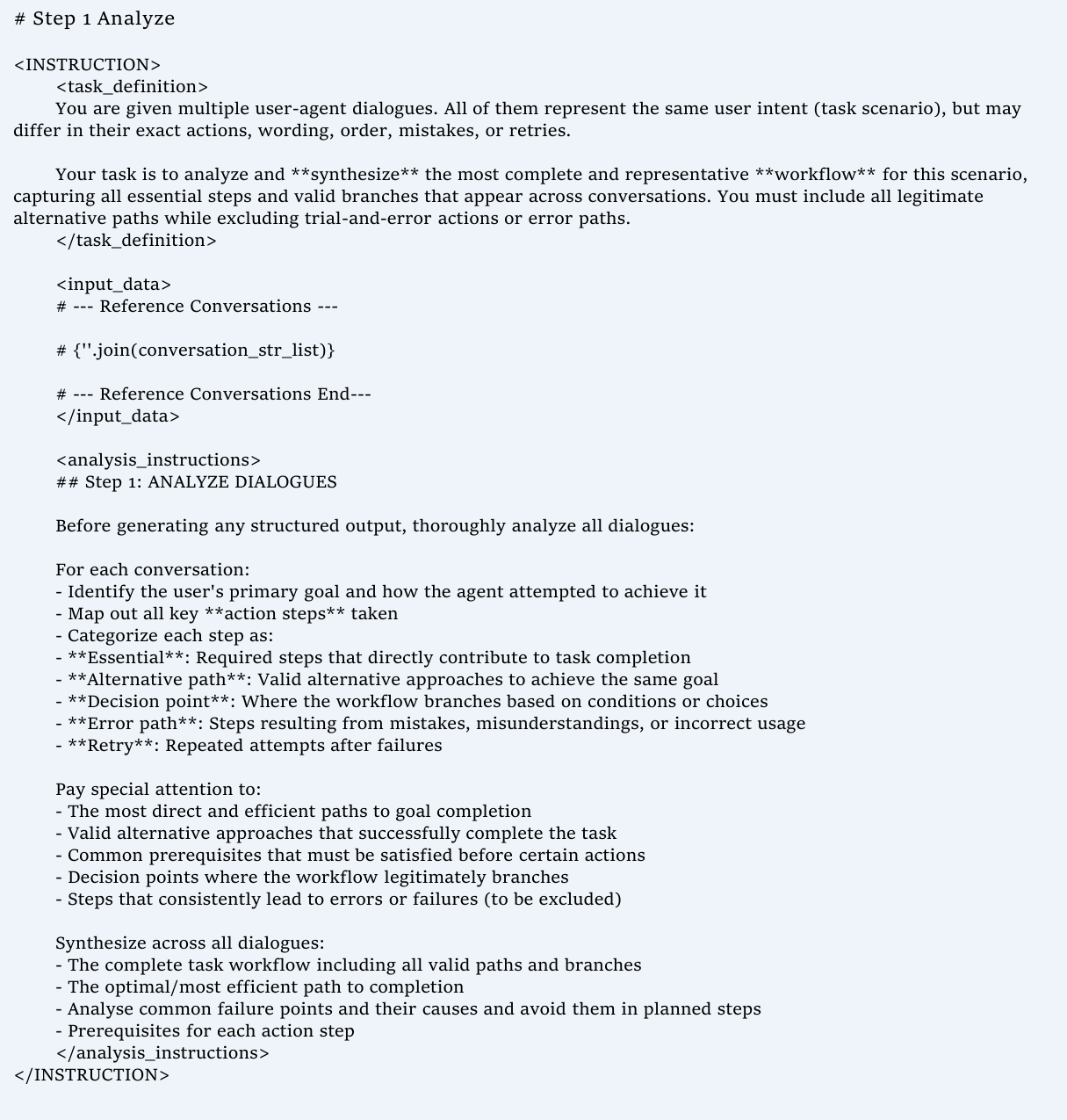}
    \caption{Analyze}
  \end{subfigure}\hfill
  \begin{subfigure}{0.49\linewidth}
    \centering
    \includegraphics[page=2,width=\linewidth]{workflow_induction.pdf}
    \caption{Draft}
  \end{subfigure}

  \vspace{0.6em}

  \begin{subfigure}{0.49\linewidth}
    \centering
    \includegraphics[page=3,width=\linewidth]{workflow_induction.pdf}
    \caption{Reflect and revise}
  \end{subfigure}\hfill
  \begin{subfigure}{0.49\linewidth}
    \centering
    \includegraphics[page=4,width=\linewidth]{workflow_induction.pdf}
    \caption{Workflow Format}
     \label{fig:workflow_induction_d}
  \end{subfigure}

  \caption{Offline Workflow Induction.}
  \label{fig:workflow_induction}
\end{figure*}

\begin{figure*}[ht!]
    \centering
    \includegraphics[width=1\linewidth]{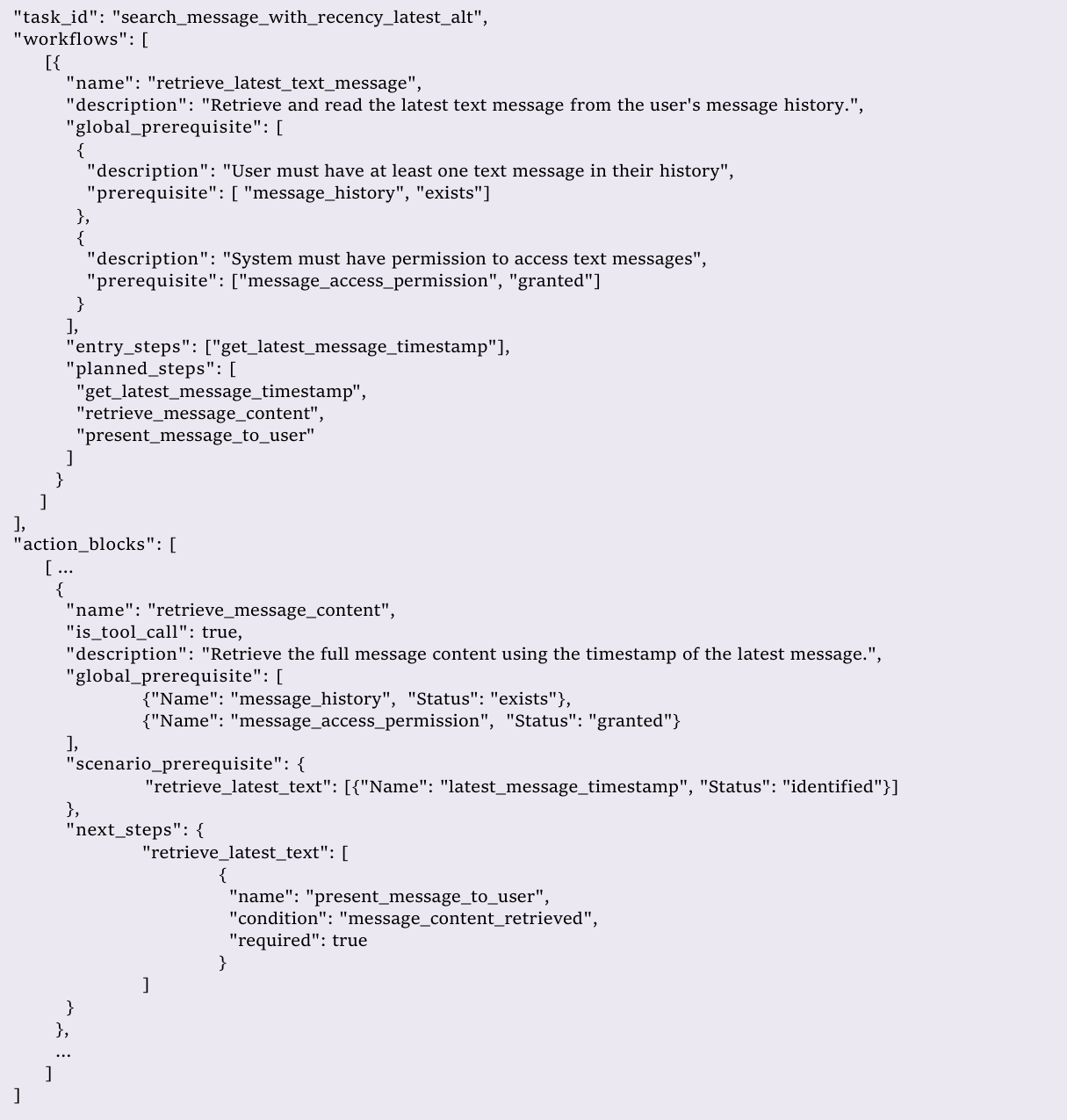}
     \caption{A workflow sample generated by WISE-Flow during offline workflow induction on ToolSandbox using Claude 3.7 Sonnet.}
    \label{fig:workflow_sample}
\end{figure*}

\paragraph{Offline Workflow Induction.} Figure~\ref{fig:workflow_induction} shows the prompts for workflow induction and the workflow format. The model used for workflow induction is the same as the agent's base model. We show a sample of induced workflow in Figure~\ref{fig:workflow_sample}.

\paragraph{Workflow Embedding and Index Construction.} We use the Cohere.embed-english-v3~\cite{cohere_embed_v3} embedding model  ("cohere.embed-english-v3") and build a FAISS~\cite{douze2025faiss} index for similarity search. Specifically, we embed each induced workflow as a rich textual document that includes its description, actions, and prerequisites. Then we index these embeddings with a cosine-similarity FAISS backend. We also build an action-level index over all action blocks. At inference time, the agent embeds the current task query and retrieves top-$K$ workflows and corresponding actions, then maps retrieved vectors back to structured workflow and actions via stored metadata for workflow-guided control.

\begin{figure*}[ht!]
    \centering
    \includegraphics[width=1\linewidth]{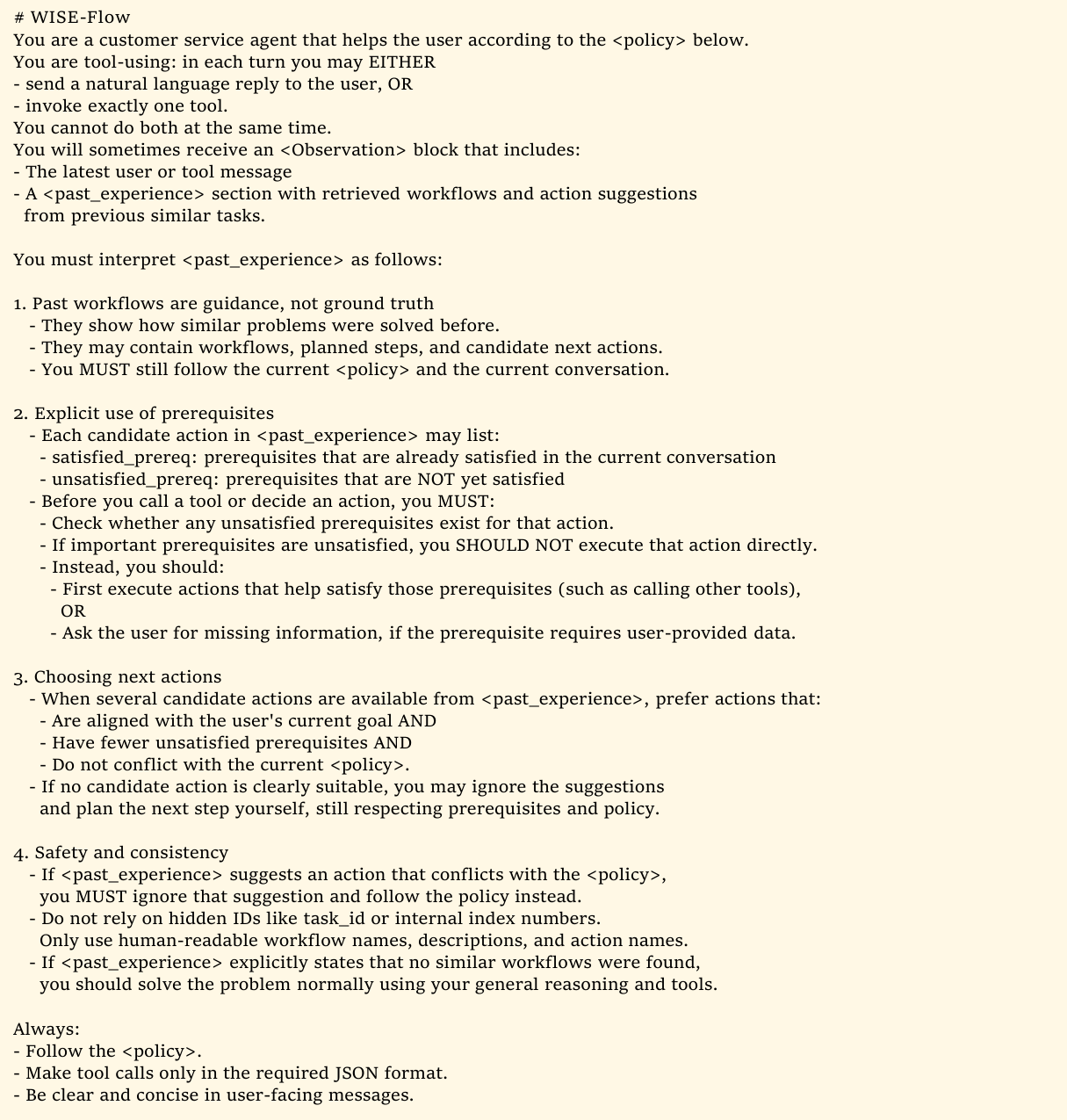}
    \caption{The system Prompt for WISE-Flow deployment on $\tau^2$-bench.}
    \label{fig:prompt-wiseflow-tau2}
\end{figure*}

\begin{figure*}[ht!]
    \centering
    \includegraphics[width=1\linewidth]{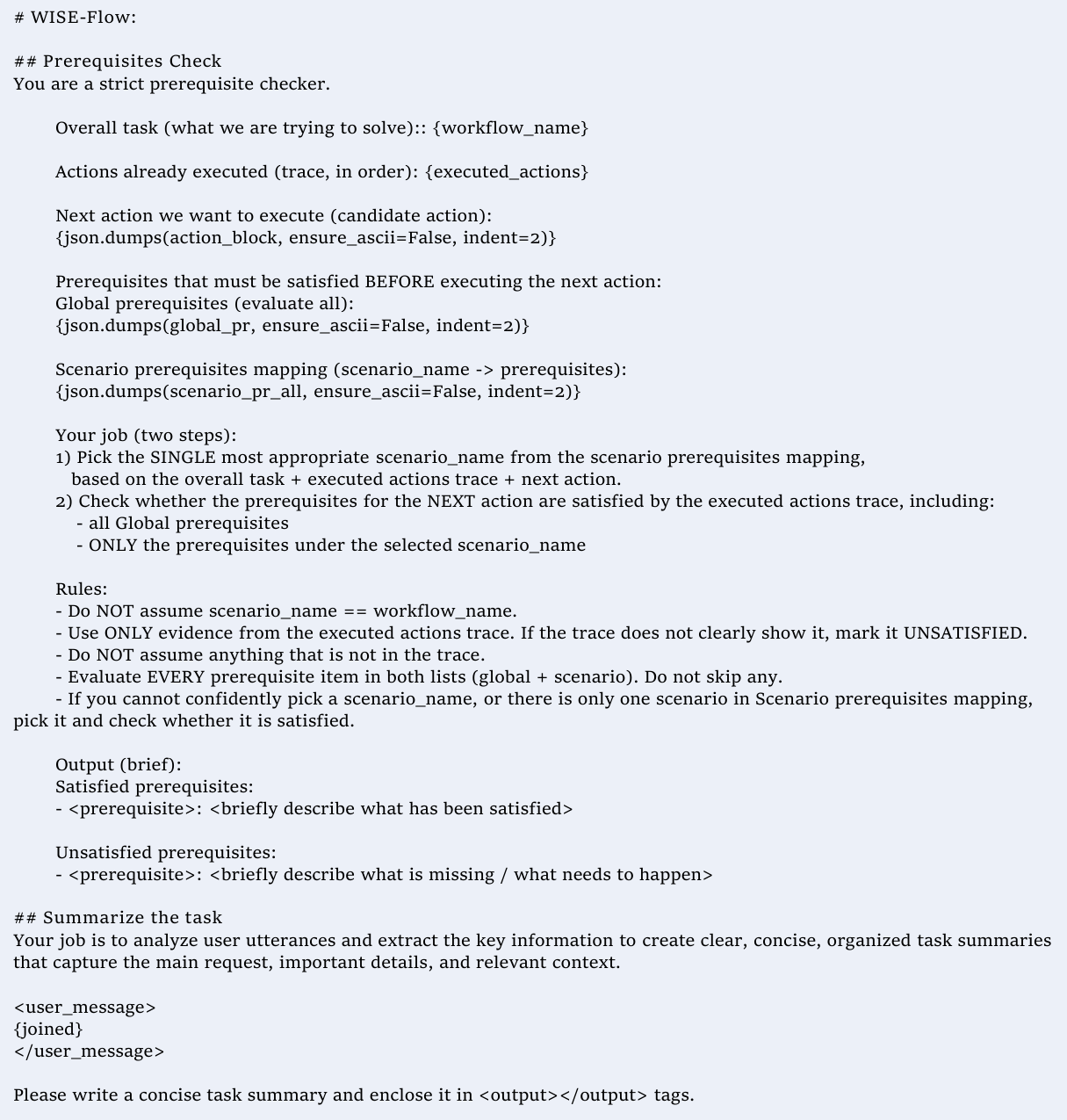}
    \caption{During WISE-Flow online deployment, we use an LLM (the same base model as the agent) to summarize the task from the current dialogue and to check whether the prerequisites of the retrieved actions are satisfied. The figure shows the two prompts used for task summarization and prerequisite checking.}
    \label{fig:prompt-wiseflow-sum-precheck}
\end{figure*}

\paragraph{Online Workflow-guided Agent Control.} For online deployment, we use the same system prompt as Reflexion and REMEMBER on ToolSandbox, as shown in Figure~\ref{fig:prompt_baseline_toolsandbox}, and use the prompt shown in Figure~\ref{fig:prompt-wiseflow-tau2} on $\tau^2$-bench. We also introduce an LLM-based pre-processing step for online retrieval. As shown in Figure~\ref{fig:prompt-wiseflow-sum-precheck}, we prompt the same base model as the agent to (i) generate a task summary from the current dialogue as the retrieval query for workflows and (ii) after progress alignment, check whether the prerequisites of the retrieved action blocks are satisfied.

\begin{table*}[]
\centering
\begin{tabular}{|c|c|c|c|c|c|}
\cline{1-6}
Model                                               & Method     & \multicolumn{1}{c|}{Similarity $\uparrow$} & \multicolumn{1}{c|}{$F_\beta\uparrow$} & \multicolumn{1}{c|}{MMR $\downarrow$} & \multicolumn{1}{c|}{SR$\uparrow$}  \\ \cline{1-6}
\multirow{6}{*}{Qwen3-235B-A22B}                              & LLM-only   & 0.0647 & 0.0652                       & 5.96\%                                        & 12.27\%                                \\ \cline{2-6}
                                                    & ReAct      & 0.0661                           & 0.0628                       & 6.32\%                                        & 13.38\%                               \\ \cline{2-6}
                                                    & Reflexion  & 0.1428                           & 0.1439                       & 13.88\%                                        & 19.60\%                              \\ \cline{2-6}
                                                    & REMEMBERER & 0.0586                           & 0.0558                       & 5.64\%                                        & 13.25\%                               \\ \cline{2-6}
                                                    & AutoGuide  &  0.0359                           &  0.0327                       &  3.31\%                                        &  9.70\%                               \\ \cline{2-6}
                                                    & WISE-Flow  &  0.0445                            &  0.0382                        &  3.81\%                                         &  10.46\%                                \\ \cline{1-6}
\multicolumn{1}{|c|}{\multirow{6}{*}{Claude 3.7 Sonnet}} & LLM-only   & 0.0339                           & 0.0322                       & 3.20\%                                        & 5.01\%                                \\ \cline{2-6}
\multicolumn{1}{|c|}{}                              & ReAct      &  0.0445                           & 0.0413                       & 4.15\%                                        & 8.74\%                                \\ \cline{2-6}
\multicolumn{1}{|c|}{}                              & Reflexion  & 0.0766                           & 0.0755                       & 7.64\%                                        & 16.12\%                                \\ \cline{2-6}
\multicolumn{1}{|c|}{}                              & REMEMBERER &  0.0228                           &  0.0212                       &  2.09\%                                        &  5.02\%                                 \\ \cline{2-6}
\multicolumn{1}{|c|}{}                              & AutoGuide  &  0.0196                           & 0.0131                       & 1.32\%                                        & 3.58\%                                \\ \cline{2-6}
\multicolumn{1}{|c|}{}                              & WISE-Flow  & 0.0353                           &  0.0334                       &  3.36\%                                         & 7.68\%                               \\ \cline{1-6}
\end{tabular}
\caption{Standard deviation of the results in Table~\ref{tab:all_toolsandbox}. }
\label{tab:all_toolsandbox_std}
\end{table*}

\begin{table}[]\small
\centering
\begin{tabular}{|c|l|c|c|c|}
\hline
Model                                                    & \multicolumn{1}{c|}{Method} & pass\textasciicircum{}1 & $F_\beta$ & MMR    \\ \hline
\multirow{6}{*}{\makecell[l]{Qwen3-\\235B-\\A22B}}                         & LLM                    & 0.2983                  & 0.2343  & 0.2367 \\ \cline{2-5} 
                                                         & ReAct                       & 0.2189                  & 0.2381  & 0.2379 \\ \cline{2-5} 
                                                         & Refl.                   & 0.2228                  & 0.1554  & 0.1551 \\ \cline{2-5} 
                                                         & REM                  & 0.2696                  & 0.2224  & 0.2215 \\ \cline{2-5} 
                                                         & AutoG                   & 0.1925                  & 0.1620  & 0.1613 \\ \cline{2-5} 
                                                         & WISE                   & 0.2482                  & 0.1447  & 0.1445 \\ \hline
\multicolumn{1}{|l|}{\multirow{6}{*}{{\makecell[l]{Claude3.7\\Sonnet}}}} & LLM                    & 0.2037                  & 0.1628  & 0.1632 \\ \cline{2-5} 
\multicolumn{1}{|l|}{}                                   & ReAct                       & 0.3283                  & 0.2541  & 0.2547 \\ \cline{2-5} 
\multicolumn{1}{|l|}{}                                   & Refl.                   & 0.1773                  & 0.1536  & 0.1528 \\ \cline{2-5} 
\multicolumn{1}{|l|}{}                                   & REM                  & 0.1722                  & 0.1541  & 0.1542 \\ \cline{2-5} 
\multicolumn{1}{|l|}{}                                   & AutoG                   & 0.1519                  & 0.1430  & 0.1430 \\ \cline{2-5} 
\multicolumn{1}{|l|}{}                                   & WISE                   & 0.1671                  & 0.1216  & 0.1217 \\ \hline
\end{tabular}
\caption{Standard deviation of the results in Table~\ref{tab:all_tau2}. (LLM = LLM-only, Refl. = Reflexion, REM = REMEMBERER, AutoG = AutoGuide, WISE = WISE-Flow.) }
\label{tab:all_tau2_std}
\end{table}

\subsection{Experiment Results}
\subsubsection{Effect of Multi-Pass Workflow Induction}
\label{sec:exp_multi-pass}
We examine the effect of the workflow induction procedure on representation quality, i.e., whether the multi-pass induction of WISE-Flow produces workflows that are more reliable, where we explicitly separate workflow induction into three stages (analyse, draft, and reflect \& revise). We compare it with a one-pass setting, where the model completes all stages with a single instruction.

As shown in Table~\ref{tab:effect-of-3pass}, three-pass induction outperforms one-pass induction across all metrics. While it improves the overall success rate by about $3\%$, the most striking effect is on one-shot success: the one-shot success rate increases by about $20\%$, and the trial-and-error success rate (i.e., success achieved only after intermediate failures) drops sharply by $17.5\%$. Together, these results show that three-pass induction does not merely raise final success, but substantially increases the likelihood of succeeding on the first attempt. This suggests that the proposed three-pass procedure induces higher-quality workflows that are more accurate and better aligned with the prior successful experiences, allowing the agent to solve tasks without relying on iterative trial-and-error.

\begin{table}[h!]
\centering
\begin{tabular}{|l|c|c|}
\hline
                          & \makecell[c]{One-Pass\\ (Single-shot)}   & \makecell[c]{Three-Pass}    \\ \hline
Similarity $\uparrow$               & 0.8659                 & $ {\mathbf{0.8824}}$                                       \\ \hline
$F_\beta $ $\uparrow$                  & 0.9256                 & $ {\mathbf{0.9433}}$                                       \\ \hline
MMR $\downarrow$ & 7.44\%                 & $ {\mathbf{5.77\%}}$                                       \\ \hline
SR $\uparrow$                       & 84.63\%                & $ {\mathbf{87.50\%}}$                                      \\ \hline
\makecell[l]{SR\\ in one-shot $\uparrow$}             & 65.88\%                & $ {\mathbf{86.25\%}}$                                      \\ \hline
\makecell[l]{SR with \\ trial-and-error $\downarrow$}   & 18.75\%                & $ {\mathbf{1.25\%}}$                                       \\ \hline
\end{tabular}
\caption{Comparison of one-pass (single-shot) and three-pass (analyze--draft--reflect\&revise) workflow induction. Experiments are conducted on ToolSandbox with Claude 3.7 Sonnet as the base model, using trajectory-wise ground-truth-based retrieval. For each test task, we retrieve all workflows induced from all trajectories of the same task, eliminating retrieval randomness and workflow aggregation noise, and isolating the effect of the induction procedure.}
\label{tab:effect-of-3pass}
\end{table}

\subsubsection{Investigation of Generalization Across Tasks}
\label{sec:exp_cross_task}
To assess whether workflows induced by WISE-Flow generalize across tasks, we evaluate cross-task reuse of retrieved workflows. As shown in Section~\ref{sec: overall-dyn}, mixing experience from multiple tasks can be beneficial. In this experiment, agents retrieve workflows from a pool of experiences collected from other tasks, where we explicitly exclude the experience from the same task. We evaluate whether cross-task borrowing can still improve performance. This experiment aims to test the transferability and practical value of structured workflows beyond within-task reuse.

\begin{figure}[h!]
    \centering
    \includegraphics[width=1\linewidth]{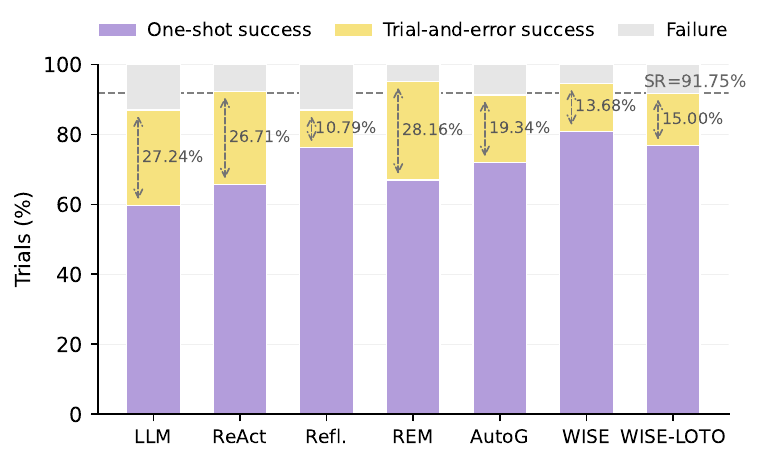}
    
    \caption{Evaluation of generalization across tasks on ToolSandbox. We compare the outcome composition of different methods, decomposing trials into one-shot success, trial-and-error success, and failure. WISE-Flow leave-one-task-out (WISE-LOTO) excludes the same-task experience from the retrieval pool.}

    \label{fig:toolsandbox-claude-transfer}
\end{figure}

From Figure~\ref{fig:toolsandbox-claude-transfer}, we observe that WISE-Flow remains competitive even when retrieving workflows only from other tasks. Although excluding same-task experience leads to a clear drop compared with standard WISE-Flow, WISE-Flow (LOTO) still outperforms several baselines in both overall success rate and one-shot success rate, such as Reflexion and AutoGuide. This suggests that the structured workflows induced by WISE-Flow capture reusable patterns that can transfer across tasks.

\subsubsection{Effectiveness of WISE-Flow Retrieval}
\label{sec:exp_retrieval_effect}
To verify that WISE-Flow benefits from retrieving relevant workflows, rather than arbitrarily selecting from the offline-induced workflow pool, we evaluate a variant of the WISE-Flow agent that randomly retrieves workflows from the same pool and compare it with the standard WISE-Flow which uses similarity-based retrieval. As shown in Table~\ref{tab:ablation_random_retrieve}, the random-retrieval variant exhibits consistent degradations across all metrics, with the overall success rate decreasing by $3\%$ and the one-shot success rate (i.e., success without any intermediate failures) dropping by $15\%$. These results indicate that WISE-Flow's gains are driven by retrieval quality. The agent retrieves workflows that are genuinely relevant for the query, whereas randomly selected workflows cannot achieve comparable performance.

\begin{table}[h!]
\centering
\begin{tabular}{|l|c|c|}
\hline
Retrieval                & Similarity-Based & Random  \\ \hline
Similarity $\uparrow$     & $ {\mathbf{0.9099}}$           & 0.9044  \\ \hline
F\_beta $\uparrow$        & $ {\mathbf{0.9745}}$           & 0.9660  \\ \hline
MMR $\downarrow$          & $ {\mathbf{2.46\%}}$           & 3.21\%  \\ \hline
SR $\uparrow$             & $ {\mathbf{94.61\%}}$          & 91.45\% \\ \hline
SR in one-shot $\uparrow$ & $ {\mathbf{80.92\%}}$          & 65.79\% \\ \hline
\end{tabular}
\caption{Performance comparison of WISE-Flow using standard similarity-based retrieval versus random retrieval on ToolSandbox.}
\label{tab:ablation_random_retrieve}
\end{table}

\subsection{Misc}
We verified that our use of third-party artifacts was consistent with their stated intended use whenever such guidance was provided. In particular, we used existing models, benchmarks, and datasets solely for their intended purpose and research purpose, and we followed any licensing and usage constraints (including platform terms for hosted models). We did not use these artifacts in ways that contradict their documented scope.

For artifacts we create, we specify the intended use as research evaluation and reproducibility. We ensure compatibility with the original access conditions of upstream resources: we do not redistribute proprietary model artifacts or restricted third-party materials, and we avoid releasing raw dataset contents or any derivatives that would violate upstream terms. If any data were accessed under research-only conditions, any derived outputs are used and shared only within research contexts, consistent with those conditions.

\subsubsection{License}
All proprietary models used in our experiments are used under the AWS Customer Agreement and AWS Service Terms; where applicable, Amazon Bedrock serverless third-party models are additionally subject to the relevant third-party model EULA/terms listed by AWS. Specifically, Qwen3-235B-A22B uses Apache License 2.0, Claude 3.7 Sonnet is Proprietary (Anthropic Terms of Service), and cohere.embed-english-v3 is Proprietary (Cohere / AWS Bedrock Terms of Use).

Meanwhile, the datasets and the benchmarks used are licensed as follows: (1) $\tau^2$-bench: MIT License; (2) ToolSandbox: Apple Software License (\url{https://github.com/apple/ToolSandbox/blob/main/LICENSE}).

\subsubsection{Model Usage}
Qwen3-235B-A22B is a MoE model with 235B total parameters. For Claude 3.7 Sonnet and cohere.embed-english-v3, the providers do not publicly disclose parameter counts in their documentation, so we report the model identifiers.

Because inference was performed via AWS Bedrock APIs and we did not log aggregate usage (e.g., total requests/tokens/cost), we cannot report a precise total computational budget (GPU-hours). The underlying accelerator hardware for Bedrock-hosted inference is not exposed to users.

% This is an appendix.

\end{document}